\documentclass[11pt]{article}
\usepackage{fullpage}
\usepackage{times}
\usepackage{authblk}
\usepackage{graphicx}
\usepackage{amsmath, amssymb}
\usepackage{tabularx}
\usepackage{array}
\usepackage[hidelinks]{hyperref}
\title{\textbf{Hybrid Data can Enhance the Utility of Synthetic Data for Training Anti-Money Laundering Models}}

\author[1]{\textbf{Rachel Chung}}
\author[2]{\textbf{Pratyush Nidhi Sharma}}
\author[2]{\textbf{Mikko Siponen}}
\author[3]{\textbf{Rohit Vadodaria}}
\author[1]{\textbf{Luke Smith}}
\affil[1]{College of William and Mary\\ 
\textnormal{rachel.chung@mason.wm.edu}, \textnormal{ltsmith02@wm.edu}}
\affil[2]{The University of Alabama\\ 
\textnormal{pnsharma@ua.edu}, \textnormal{tmsiponen@ua.edu}}
\affil[3]{CGI Federal\\ 
\textnormal{rohitv2019@gmail.com}}
\date{September, 2025}

\begin{document}
\maketitle

\begin{abstract}
Money laundering is a critical global issue for financial institutions. Automated Anti-money laundering (AML) models, like Graph Neural Networks (GNN), can be trained to identify illicit transactions in real time. A major issue for developing such models is the lack of access to training data due to privacy and confidentiality concerns. Synthetically generated data that mimics the statistical properties of real data but preserves privacy and confidentiality has been proposed as a solution. However, training AML models on purely synthetic datasets presents its own set of challenges. This article proposes the use of hybrid datasets to augment the utility of synthetic datasets by incorporating publicly available, easily accessible, and real-world features. These additions demonstrate that hybrid datasets not only preserve privacy but also improve model utility, offering a practical pathway for financial institutions to enhance AML systems.
\end{abstract}


\section{Introduction}
Money laundering involves legitimizing illicit funds through bank accounts and businesses to hide their origins. The amount of money laundered annually is estimated to be 2--5\% of the global GDP, or \$800 billion--\$2 trillion in current U.S. dollars.\footnote{\url{https://www.unodc.org/unodc/en/money-laundering/overview.html}} The situation is getting worse. The number of money laundering incidents within the U.S. alone has increased by 14.3\% since 2019 and the median loss per incident has increased from \$257,200 in 2019 to \$554,353 by 2023. \footnote{\url{  https://www.ussc.gov/research/quick-facts/money-laundering}}  Even advanced nations do not have sufficient capabilities to address this worsening issue [1]. 

Money laundering poses a critical threat to financial systems globally. U.S. law mandates that all financial institutions must have anti-money laundering procedures in place and are liable to identify and prevent money laundering activities. Per the Patriot Act, legislators have instilled extensive responsibility upon accountants and auditors to prevent and detect money laundering [1].

Artificial Intelligence models are promising solutions for automated detection of money laundering activities. While real-world datasets have been used for training AML models [2], such applications are rare since access to real datasets is generally prohibited due to privacy concerns and government regulations [3].  

Synthetic data is an increasingly common solution to overcome such challenges with real datasets [3], [4], [5], [6]. Recent work has heavily focused on algorithmically refining synthetic data generation methods [7], [8]. However, model training on purely synthetic datasets presents its own set of challenges [9]. We show that the utility of synthetic AML datasets can be augmented by incorporating real-world and publicly available features.

This paper makes three contributions. First, we introduce a hybrid data approach that combines synthetic transaction data with public country-level indicators, preserving privacy while enhancing realism. Second, we provide an empirical demonstration that such hybridization significantly improves the predictive utility of graph neural networks for AML detection. Third, we highlight implications for practitioners and policymakers, showing how publicly available contextual features can strengthen compliance tools without risking sensitive data leakage.

\section{Background}
Synthetic data is artificially created to mimic a real dataset by learning its statistical properties and patterns through generative algorithms. Synthetic data has gained increasing attention in the financial domain as a means of addressing the scarcity and sensitivity of real-world datasets. Recent research  have demonstrated the potential of synthetic financial data for model training while preserving privacy [3], [4], [6]. For instance, Altman et al. [4] developed realistic synthetic financial transactions to facilitate AML model testing, while Borrajo et al. [5] employed agent-based simulation to generate adversarial environments for money laundering detection. Recent advances have focused on improving the fidelity of synthetic data generation using approaches such as graph-based simulation [7] and diffusion models [8]. Nevertheless, prior work also shows the limitations of synthetic data, particularly its inability to fully capture the complexity of real-world transaction contexts [3], [9].

In recent years, graph-based anomaly detection methods have emerged as powerful tools for fraud and AML detection. Graph neural networks (GNNs) can model heterogeneous relationships between entities and transactions, outperforming classical machine learning approaches in imbalanced anomaly detection tasks [2], [12]. Applications of GNNs in financial crime include Ethereum fraud detection [13] and self-supervised representation learning for money laundering [14]. Johannessen and Jullum [2] showed that heterogeneous GNNs could effectively identify money launderers by leveraging relational structures, while Pourhabibi et al. [12] provided a systematic review highlighting the promise of graph-based approaches for fraud detection more broadly. These studies provide evidence for the importance of network structure in capturing hidden patterns of illicit behavior.

Despite these advancements, little attention has been paid to integrating synthetic data approaches with graph-based AML detection methods. Synthetic data generation and the use of GNNs in AML tasks have largely evolved as separate lines of inquiry. It remains unclear how well synthetic datasets perform with GNNs in AML detection, or whether incorporating real-world features can enhance their effectiveness.  Our work contributes to bridging this gap by demonstrating how hybrid datasets—constructed by augmenting synthetic transactions with public, country-level contextual features—can improve the realism and predictive utility of GNN-based AML detection. By doing so, we extend prior efforts on synthetic data and graph modeling, offering a new pathway to enhance AML detection under strict privacy constraints.

\subsection{The Role of Synthetic Data in Developing AML Models}
Given the confidentiality and data availability concerns, organizations including JP Morgan, IBM, and Society for Worldwide Interbank Financial Telecommunication (Swift) have produced synthetic AML datasets for model training purposes [4], [5], [10]. In addition to addressing privacy concerns, synthetic data is used when there might be class imbalance where one outcome occurs much more frequently than the other, such as in the case of money laundering transactions.

While synthetic datasets are useful for model training, they have limitations. One concern is how realistic the synthetic data is and whether it can capture the complexity of real-world financial systems [3]. In regulated environments, such as the financial sector, very few features may be available, even in a synthetic dataset. For example, the JP Morgan’s AML dataset [5] lacks features about the transactions (i.e., edge features), such as the method by which the transaction was made – online, in-person, mobile app, and the nodes (i.e., node features), such as age and business type of an account [11]. The dataset also lacks features that could provide contextual information for cross-border transactions. If included in the synthetic dataset, such features could greatly improve model training.

\subsection{Model Performance Issues with Synthetic AML Dataset}
Graph Neural Networks (GNNs) offer a promising approach for money laundering detection by accounting for both transaction-specific characteristics and the networked connections among transactions [2]. The use of graphs to model monetary transactions between financial entities has become increasingly popular, particularly for fraud detection [12]. GNNs have outperformed classical machine learning methods, such as support vector machines, by achieving higher predictive accuracy for anomaly detection with unbalanced data [2], [13], [14].

Following [2], we implemented a Relational Graph Convolution Network (RGCN) using the PyTorch library, and trained the model on JPMorgan’s synthetic AML dataset [5]. This dataset comprises individual labeled transactions generated using an agent-based simulator, which observes behavioral traces to produce realistic and novel transaction records [5]. The method simulates the country of origin and destination for each transaction, allowing the current study to incorporate real-world measures at the country level. 

The dataset includes 523,877 individual transactions from 16 countries. The transaction labels (either “GOOD” or “BAD”) are imbalanced, with only 20\% labeled as “BAD”, i.e., fraudulent. Simulated features of each account include ID, financial institution, and the country. Each transaction includes type and dollar value. Transaction values have a mean of \$148,339.46, and a standard deviation of \$473,121.20. 

Our RGCN model involved two convolution layers, a hidden channel of 16 neurons, and outputs a binary classification where 1 indicates “BAD” (suspicious) and 0 indicates “GOOD” (non-suspicious) transactions. The model achieved an accuracy of nearly 65\%, meaning that it correctly classified the majority of instances. However, the F1-score was only 7.75\%. This highlights a significant imbalance between precision (the proportion of true positives among all positive predictions) and recall (the proportion of true positives among all actual positives). The low F1-score indicates the model struggled to balance identifying true positives and avoiding false positives. Finally, the AUC (Area Under the Curve) measures the model's ability to distinguish between the classes, with higher values indicating better performance. Our model’s AUC value of 43.64\% indicates that it performed worse than a coin flip! Overall, the model performed poorly in distinguishing between “BAD” and “GOOD” instances.

\section{Hybrid Data: Augmenting Synthetic Data with Real-world Features}
In light of the underwhelming model performance when trained on the synthetic dataset, we decided to augment our feature set with real-world publicly available country-level information for two reasons. First, a concern about using synthetic data to train AI models is the extent to which the synthetic data is realistic [3], [15]. Incorporating real-world country-level features may enhance the dataset's realism by providing important contextual cues (country-specific risks such as regulatory and legal environment, financial infrastructure etc.) for improving model performance, especially for cross-border transactions [10]. Second, the easily accessible and publicly available nature of country-level data makes this method practical without running into privacy or confidentiality issues. 

Researchers have recommended incorporating additional features to enhance model performance [13]. However, while prior work has utilized the origin and destination country as a specific account feature [14], features providing important country-level contextual cues have been missing. Thus, we collected the following four country-level metrics from year 2022, and combined them with the synthetic dataset to create our hybrid dataset:

1.	Basel Anti-Money Laundering Index by the Basel Institute on Governance is a quantitative tool that assesses the risk of money laundering in countries. It considers legal and political frameworks, financial regulations, and corruption levels to provide a comprehensive evaluation of a country's susceptibility to illicit activities.\footnote{\url{https://baselgovernance.org/basel-aml-index}}

2.	Digital Evolution Index by Tufts University measures the extent to which a country has embraced digital technologies and integrated them into its economy. Factors include internet penetration, digital infrastructure, e-governance initiatives, and the overall digital literacy of the population. A high Digital Evolution Index suggests a more digitally advanced and connected society.\footnote{\url{https://digitalevolutionindex.tufts.edu/trajectory}}

3.	Corruption Perceptions Index (CPI) by Transparency International is an indicator of perceived corruption levels in the public sector of countries worldwide. Countries are ranked on a scale from highly corrupt to very clean, providing insights into the perceived integrity of public institutions and governance.\footnote{\url{https://www.transparency.org/en/cpi/}}   

4.	Gross Domestic Product (GDP) per Capita is a key economic indicator that represents the average economic output per person in each country. 

\subsection{Model Performance with Hybrid Dataset}

Our RGCN model performed much better when the synthetic dataset was augmented with real-world country-level features. Specifically, accuracy rate improved from 64.93\% to 83.39\%, F1-score improved from 7.75\% to 59.37\%, and AUC improved from 43.64\% to 74.63\%. This provides evidence that, at least in the case of AML detection, augmenting the synthetic dataset with real-world country-level features can significantly improvement in the model’s predictive accuracy. 

\section{Discussion}

We evaluated the effectiveness of a hybrid dataset for money-laundering detection model training. Using a RGCN model, we were able to improve money laundering detection performance with the hybrid dataset. The evidence suggests that, at least in some situations, augmenting synthetic datasets with real-world features can be a useful method for improving the performance of money laundering detection models. Table 1 provides a comparison of synthetic and hybrid AML datasets. 

\begin{table}[ht]
\centering
\caption{Synthetic vs. Hybrid AML Datasets}
\label{tab:synthetic_hybrid}
\renewcommand{\arraystretch}{1.1}
\small
\begin{tabularx}{\columnwidth}{>{\raggedright\arraybackslash}p{2.2cm}X X}
\hline
 & \textbf{Synthetic Dataset} & \textbf{Hybrid Dataset} \\
\hline
\textbf{Definition} & 
Artificially created dataset to mimic a real dataset by learning its statistical properties and patterns through generative algorithms &
Dataset that combines synthetically generated data with real-world features \\[4pt]

\textbf{Realism} & 
Limited realism, as dataset is typically focused on simulating transaction flows &
Enhanced realism via additional real-world features such as relevant context for source and destination countries \\[4pt]

\textbf{Privacy} & 
Excellent privacy as data has no ties to real-world entities &
Excellent privacy as long as the real-world features do not contain privacy sensitive information \\[4pt]

\textbf{Class Imbalance} & 
Dataset can be artificially balanced to address class imbalance challenges in model training &
The synthetic portion of the dataset can be artificially balanced to address class imbalance challenges in model training \\
\hline
\end{tabularx}
\end{table}

Our findings have several implications. The incorporation of real-world features presents an incremental step towards alleviating growing concerns about model training on purely synthetic data [9]. Moreover, the use of country-level features may be particularly relevant for modeling cross-border transactions, such as those handled by Swift [10]. More importantly, financial institutions can significantly improve their AML detection methods by utilizing GNN models trained on hybrid datasets that integrate easily and publicly accessible country-level attributes with synthetic datasets.

Our work focused primarily on country-level information as node features. While this is one way to improve model performance there are other steps that can be taken. In particular, we only focused on four country-level features to highlight how AML detection may be improved using easily available public information. Addition of other such features may further improve model performance. Future work can explore transaction (i.e., edge) features [2], in addition to such node features, if more transaction features become available in synthetic datasets.  

One can also explore alternative functions of aggregation, in addition to summation implemented here. Architecturally, one can explore the value of implementing a GNN model with differing node embedding aggregations [14], and the potential benefit of incorporating other types of nodes in the model. Given that transaction data and country-level features are at different levels of abstraction, it is also promising to consider GNN architectures that sequentially integrate different orders of information. Finally, with the rising success of generative AI, such as GPT models, there are also many opportunities to explore unstructured synthetic data, such as generated text, PDF documents, images, and audio files for training money laundering detection models. 

Prior studies on synthetic data for finance have primarily emphasized generation fidelity [3], [4], [6], [7], [8], while graph-based approaches have focused on modeling network structures to improve fraud and AML detection [2], [12], [14]. These areas of research have largely progressed in parallel, with little integration between them. Our study addresses this gap by showing that hybrid datasets, created by combining synthetic transactions with public country level indicators, can significantly enhance the predictive performance of graph neural networks. In doing so, we extend prior work that has treated synthetic data and graph-based modeling as separate challenges, showing instead that their combination offers a practical and privacy-preserving pathway for advancing AML detection.

While there is understandable current interest in refining algorithms for synthetic data generation, we believe the development of robust AML models will require a combination of approaches. These include multi-level feature integration, leveraging both structured and unstructured data, and employing advanced GNN architectures. By blending these techniques, future AML models will be better equipped to adapt to the evolving complexity of money laundering schemes and provide financial institutions with more comprehensive detection tools. Hybrid AML data can play an important role in driving this evolution.    

\section{Conclusion}
Hybrid datasets that combine privacy-preserving synthetic transactions with public, country-level indicators provide a pragmatic path to more realistic and effective AML model training. By aligning training data more closely with deployment-time context while maintaining strict privacy guarantees, institutions can improve detection performance and strengthen compliance workflows without accessing sensitive customer data.

\end{document}